\documentclass{article}

\PassOptionsToPackage{numbers, compress}{natbib}

\usepackage[preprint]{neurips_2024}

\usepackage[utf8]{inputenc} 
\usepackage[T1]{fontenc}    
\usepackage{hyperref}       
\usepackage{url}            
\usepackage{booktabs}       
\usepackage{amsfonts}       
\usepackage{nicefrac}       
\usepackage{microtype}      
\usepackage{xcolor}         
\usepackage{graphicx}
\usepackage{amsmath}
\usepackage{amssymb}
\usepackage{mathtools}
\usepackage{amsthm}
\usepackage{algorithm}
\usepackage{algpseudocode}
\usepackage{changepage}
\usepackage{sidecap}
\usepackage{bbm}
\usepackage{tikz}
\usepackage{subcaption}
\usepackage{commath}
\usepackage{xcolor,colortbl}

\definecolor{baselinecolor}{gray}{.9}
\newcommand{\baseline}[1]{\cellcolor{baselinecolor}{#1}}

\newlength\savewidth

\theoremstyle{plain}

\theoremstyle{definition}

\theoremstyle{remark}

\definecolor{mydarkblue}{rgb}{0,0.08,0.45}
\hypersetup{ %
    colorlinks=true,
    linkcolor=mydarkblue,
    citecolor=mydarkblue,
    filecolor=mydarkblue,
    urlcolor=mydarkblue,
}

\title{Unified Auto-Encoding with Masked Diffusion}

\author{
    Philippe Hansen-Estruch$^1$ ~
    Sriram Vishwanath$^1$ ~ 
    Amy Zhang$^1$ ~ 
    Manan Tomar$^2$ ~\\
    University of Texas at Austin$^1$, University of Alberta$^2$ \\
    \texttt{philippehansen@utexas.edu}
}

\begin{document}

\maketitle

\begin{abstract}
  At the core of both successful generative and self-supervised representation learning models there is a reconstruction objective that incorporates some form of image corruption. Diffusion models implement this approach through a scheduled Gaussian corruption process, while masked auto-encoder models do so by masking patches of the image. Despite their different approaches, the underlying similarity in their methodologies suggests a promising avenue for an auto-encoder capable of both de-noising tasks. We propose a unified self-supervised objective, dubbed Unified Masked Diffusion (UMD), that combines patch-based and noise-based corruption techniques within a single auto-encoding framework. Specifically, UMD modifies the diffusion transformer (DiT) training process by introducing an additional noise-free, high masking representation step in the diffusion noising schedule, and utilizes a mixed masked and noised image for subsequent timesteps. By integrating features useful for diffusion modeling and for predicting masked patch tokens, UMD achieves strong performance in downstream generative and representation learning tasks, including linear probing and class-conditional generation. This is achieved without the need for heavy data augmentations, multiple views, or additional encoders. Furthermore, UMD improves over the computational efficiency of prior diffusion based methods in total training time. We release our code at \href{https://github.com/philippe-eecs/small-vision}{\texttt{github.com/philippe-eecs/small-vision}}.
\end{abstract}

\section{Introduction}
Conventionally, generative and representation learning objectives have two distinctly separate categories of algorithms. A long standing problem has been to bridge this gap as to have a single algorithm capable of both~\citep{li2023mage}. With the increasing performance of diffusion models as powerful generative models~\citep{rombach2022high, saharia2022photorealistic}, and self-supervised masked autoencoders as powerful representation learners~\citep{he2022masked, he2020momentum}, the need for a single, unified model that achieves both capabilities is highly desirable. Importantly, there exists strong parallels between the ideas used in diffusion models and masked auto-encoders, with the primary principle of both methods being corruption followed by reconstruction of the original input. A secondary question then arises as to what finer level design choices play the most important role in steering generative/representative capabilities.

The recent surge in popularity of diffusion models~\citep{sohl2015deep, song2019generative, ho2020denoising, song2020score} for generating high-quality image and video data has led to a renewed interest in their potential for representation learning. Notably, diffusion models have been shown to capture strong representations~\citep{preechakul2022diffusion, mukhopadhyay2023diffusion, hudson2023soda, wei2023diffusion}, rivaling those derived from contrastive learning~\citep{chen2020simple, he2020momentum} and masked autoencoder methods~\citep{he2022masked, li2023mage}. However, the such diffusion models often necessitate additional encoders and entail substantial computational demands. Moreover, in the process of optimizing for representation strength, some methods compromise the model's generative capabilities. On the other hand, MAEs have lead to robustly learning self-supervised representations, while significantly reducing the computational burden in processing the input. This is achieved by employing high masking ratios which discards much of the sequence length passed into the Vision Transformer (ViT) backbone~\citep{dosovitskiy2020image}. However, MAEs often exhibit subpar infilling and generative capabilities: producing blurry images given some context. Moreover, attempts to model the entire image sequence for better generation results, as done in MaskGIT and MAGE ~\citep{chang2022maskgit, li2023mage}, leads to a computationally intensive process unable to take advantage of the sequence length reduction as achieved by high masking in MAE.

In this paper, we look to develop a unified auto-encoder model capable of de-noising at both the token scale and at the fine grain pixel scale. Such a method can be applicable for strong representations as well as generations, while also preserving computational efficiency. We introduce Unified Masked Diffusion (UMD), which combines an asymmetric encoder-decoder architecture with a modified noise schedule derived from diffusion models. UMD adds a noise-free masked reconstruction step in addition to the standard noise schedule within the diffusion process. The noise-free step utilizes a higher masking and sampling ratio than that used for the noisy sampling steps. This simple self-supervised strategy not only aids in learning strong representations for linear-probing but also learns useful features for generation tasks, such as class-conditional generation.

Our main contributions are (1) the development of an simple self-supervised method that excels in both representation learning and generative modeling, and (2) a thorough empirical analysis of the impact of denoising autoencoders on both representation and generative quality. UMD not only competes favorably with state-of-the-art methods in terms of ImageNet linear probing accuracy and generative performance but also achieves this with fewer GPU hours required for training compared to conventional generative models. Furthermore, the simplicity of implementing UMD on top of existing MAE and DiT frameworks positions it as a highly effective alternative method.

\section{Related Work}
\label{relatedwork}

We briefly overview prior works in self-supervised representation learning, generative modeling, and diffusion models applied to self-supervised learning. 

\paragraph{Self-Supervised Representation Learning.} Self-supervised learning relies on constructing proxy tasks just from the input data, with no external labels, to learn useful representations. A prominent approach in self-supervised learning is contrastive learning \citep{oord2018representation, chen2020simple, he2020momentum, grill2020bootstrap, chen2021exploring, caron2021emerging}, which encourages representations of augmented versions of an image to be closer while pushing apart the representations of two different images. 

Another effective method for self-supervised representation learning comes from masked image modeling methods (MIM)~\citep{bao2021beit, he2022masked}. Inspired by BERT~\citep{devlin2018bert}, these methods form an objective through the reconstruction of masked parts of the input data, thereby enabling the model to capture a high level understanding of the data. The most commonly used objective is Masked Autoencoding (MAE) \citep{he2022masked}, where a significant portion of the input image is masked (75\%), hence reducing the sequence length inputted into the Vision Transformer (ViT) \citep{vaswani2017attention, dosovitskiy2020image} encoder, enabling larger batch sizes and faster training. MIM methods are tuned to learn strong representations of their data but tend to have poor and blurry reconstructions. These self-supervised methods learn effective representations of their space, but are only useful in discriminative tasks and not generative.

\paragraph{Generative Modeling.} Generative models on image spaces have advanced a lot in recent years. Generative Adversarial Networks (GANs)~\citep{goodfellow2014generative, heusel2017gans, karras2019style}, which involve training a generator to produce realistic images and a discriminator to differentiate between real and generated images, have achieved strong results in image synthesis. However, despite their ability to create high-quality images, GANs have issues with mode collapse and training instability.

Another approach is to break down image generation into two stages: tokenization of images into discrete sequences via VQVAE or VQGAN~\citep{oord2017neural, esser2021taming}, followed by auto-regressive maximum likelihood modeling~\citep{ramesh2021zero, bai2023sequential, liu2024world}. While these methods are effective for image generation, they are computationally expensive and rely on pre-trained tokenizers, adding complexity to the training process. In contrast, MaskGIT~\citep{chang2022maskgit} and MAGE~\citep{li2023mage} use a bidirectional transformer for token generation coupled with parallel decoding to expedite generation. However, these methods still require attention computations across the full sequence, missing out on the computational efficiencies of fixed masking as utilized in MAE. 

Finally, Diffusion models \citep{sohl2015deep, ho2020denoising, song2019generative}, known for their simplicity and effectiveness in image synthesis, offer a promising alternative to traditional generative approaches. Diffusion models traditionally utilized the U-Net architecture~\citep{ronneberger2015u}, however, Diffusion Transformers (DiT)~\citep{peebles2023scalable} presented a modified vision transformer (ViT) architecture~\citep{dosovitskiy2020image} that uses AdaLN conditioning~\citep{perez2018film}, achieving strong scalability and performance. While DiT requires substantial computational resources, MaskDiT~\citep{zheng2023fast} and MDT~\citep{gao2023masked} integrate MAE's masking with the diffusion noise schedule, enhancing computational efficiency. However, these methods do not explicitly test the representation capabilities of their method and, as shown in Section~\ref{ref:LinearProbeResults}, fail to achieve strong representations in downstream tasks. 

\paragraph{Representation Learning with Diffusion Models.}
Diffusion models are increasingly recognized for their representation learning capabilities. \cite{mukhopadhyay2023diffusion} and \cite{yang2023diffusion} show that the most effective representations from unconditional diffusion models are found in their intermediate layers, although these have not consistently surpassed the performance of other self-supervised methods. Extending this research,  \citep{chen2024deconstructing} further deconstructs a diffusion model (DiT) to identify the key components crucial for representation learning. This work also introduces a novel corruption scheme that enhances representation capabilities, however the method compromises the generations as a result. 

On the other hand, approaches like Diffusion-Based Representation Learning \citep{abstreiter2021diffusion} and DiffAE \citep{preechakul2022diffusion} condition diffusion models on unnoised image encodings from additional encoders, training them end-to-end. Building on this, SODA ~\citep{hudson2023soda} introduces additional views and augmentations to enhance representation performance, while DiffMAE ~\citep{wei2023diffusion} integrates masking with diffusion, reconstructing "masked" or noised patches based on the context provided by non-noised patches. Similarly, Representation-Conditioned Generation (RCG) ~\citep{li2023self} conditions a diffusion model on embeddings from a pre-trained encoder (e.g. MOCOv3~\citep{he2020momentum}) and in parallel trains a diffusion model on the representation space unconditionally, enabling un-conditional image generation. However, these methods retain the heavy training demands of traditional diffusion models and usually require additional encoder components, which complicates generating images only from simple text or class labels. 

\section{Preliminaries}
\label{prelim}
In this section, we provide a brief overview of diffusion models and masked auto-encoders. At a high level, the methods share a lot of similarity in corruption style and objective. 
\paragraph{Diffusion Models.}
Diffusion models are a class of generative models that simulate a gradual noising and denoising process through a series of latent variables. They are characterized by a Markovian forward process and a learned reverse process. Specifically, the forward process incrementally adds noise to an input image $x_0$, transitioning it through a sequence of states $x_1, \cdots, x_T$ according to a predetermined variance schedule $\beta_1, \cdots, \beta_T$. The reverse process, learned during training, aims to recover the original data from its noised version. The forward noising process is defined as:
\begin{equation}
q(x_t|x_0) = \mathcal{N}(\sqrt{\hat{\alpha}_t}x_0, (1 - \hat{\alpha}_t)I) = \sqrt{\hat{\alpha}_t}x_0 + \sqrt{(1 - \hat{\alpha}_t)}\epsilon,
\label{eq:diffusionsample}
\end{equation}

where $\hat{\alpha}_t = \prod_{s=1}^t (1 - \beta_t) \ $ and $\epsilon \sim \mathcal{N}(0, I)$ is Gaussian noise. Diffusion models are trained to estimate the reverse process, \( p_\theta(x_{t-1} \mid x_t) \), by approximating the variational lower bound of \( \int p_\theta(x_{0:T} \mid x_t) \, d(x_{0:T})
\) as computed by~\cite{sohl2015deep}. In practical implementations, this reverse process is generally conditioned on the timestep \( t \) and aims to either predict the noise \( \epsilon \) or reconstruct the original image \( x_0 \). Denoising Diffusion Probabilistic Models (DDPMs)~\citep{ho2020denoising} utilize a streamlined objective to predict the noise with network \( \epsilon_{\theta}(x_t, t)\) for a given image dataset $x_0 \sim \mathcal{D}$:
\[
\mathcal{L}_{\text{diffusion}}(\theta) = \mathbb{E}_{t \sim \mathcal{U}(1, T), \epsilon, x_0}\left[\left\| \epsilon - \epsilon_{\theta}(x_t, t)\right\|_2^2\right].
\]

Commonly the DDIM~\citep{song2020score} reverse process is used to sample the model, defined as
\begin{equation}
    x_{t-1} \leftarrow  \sqrt{\hat{\alpha}_{t-1}} \cdot \hat{x}_0 + \sqrt{1 - \hat{\alpha}_{t-1} - {\sigma_t}^2} \cdot \epsilon_{\theta}(x_t, t) + \sigma_t \cdot \epsilon,
\end{equation}

where $\sigma_t = \eta \cdot  \sqrt{\frac{1 - \hat{\alpha}_{t-1}}{1 - \hat{\alpha}_t}} \cdot \sqrt{\frac{1 - \hat{\alpha}_t}{\hat{\alpha}_{t-1}}}$, $\hat{x}_0 = \frac{x_t - \sqrt{1 - \hat{\alpha}_t} \cdot \epsilon_{\theta}(x_t, t)}{\sqrt{\hat{\alpha}_t}}$, and $\epsilon \sim \mathcal{N}(0, I)$. This sampler is particularly suited to skip over steps in the schedule which saves inference time cost. 

We also utilize classifier-free guidance~\citep{dhariwal2021diffusion, ho2022classifier} (CFG) in our generation process after fine-tuning on labels. CFG is a technique used in conditional diffusion models to improve the quality of generated samples without relying on an external classifier. In this setting, the reverse process of the diffusion model is conditioned on both the noisy image \( x_t \) and the class label \( y \), leading to \( p_\theta(x_{t-1} | x_t, y) \). This method modifies the sampling procedure to encourage the generation of samples that are highly probable given the class label \( y \) such that \( \log p(y | x_0) \) is maximized. The new sampling prediction for a given label \( y \) is:
\begin{equation}
\label{eq:CFG}
\hat{\epsilon}_\theta(x_t, t, y) = \epsilon_\theta(x_t, t, \varnothing) + s \cdot (\epsilon_\theta(x_t, t, y) - \epsilon_\theta(x_t, t, \varnothing)),
\end{equation}
where \( \epsilon_\theta(x_t, t, \varnothing) \) is the noise prediction conditioned on the null class (i.e., without the class label), and \( s > 1 \) indicates the scale of the guidance (with \( s = 1 \) recovering the standard sampling procedure). Evaluating the diffusion model with \( y = \varnothing \) is achieved by randomly dropping out \( y \) during training and replacing it with a learned "null" embedding \( \varnothing \).

\paragraph{Masked Auto-Encoders (MAE).}
The concept of Masked Auto Encoders (MAE) builds on the principles of masked training, akin to BERT~\citep{devlin2018bert} and BeIT~\citep{bao2021beit}. The core idea is to train a model to reconstruct an image from a partially masked version of it.

Given an unedited image $x_0$, MAE first employs a Vision Transformer (ViT)~\citep{dosovitskiy2020image} architecture to divide the image into a grid of patches. A subset of these patches are then randomly masked, resulting in a masked image $x^M_0$. The MAE's objective is to reconstruct the original image $p_\theta(x_0 | x^M_0)$.

One of the key features of MAE is its asymmetric encoder-decoder architecture. The encoder processes only the visible (unmasked) patches, while the decoder is tasked with reconstructing the entire image, including the masked portions. The encoder is typically a deep network, whereas the decoder is shallower. This asymmetry allows for efficient computation as the attention mechanism in the encoder operates over a reduced sequence length. This design not only enhances computational efficiency but also enables the learning of robust and rich representations of the data.

\section{A Unified Auto-Encoder Objective}
\label{acmethod}

\begin{figure*}[t]
    \centering
    \includegraphics[trim={0cm 0cm 0cm 0cm}, clip, width=0.9\linewidth]{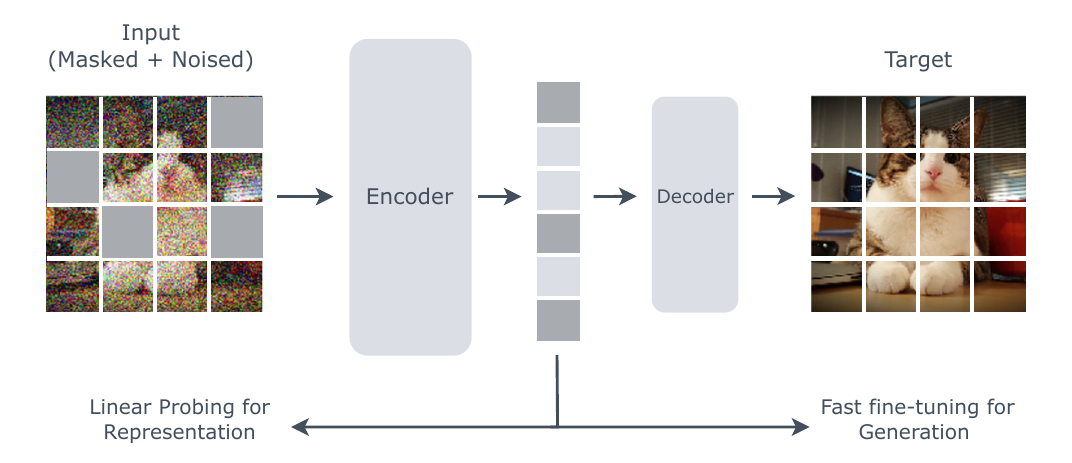}
    \vspace{2mm}
    \caption{\textbf{Unified Masked Diffusion (UMD).} UMD combines random masking with the fine-grain noise schedule used in diffusion. The target is to predict the original image. For generation, UMD finetunes the encoder and decoder on unmasked noised images + class labels. }
    \label{fig:main}
\end{figure*}

We introduce Unified Masked Diffusion (UMD), an asymmetric encoder-decoder model based on the Vision Transformer (ViT)~\citep{dosovitskiy2020image}. We initially apply Gaussian noise following a diffusion schedule to an image, feed this image as patches into the UMD encoder, and then randomly mask portions of the image (see Figure~\ref{fig:main}). Contrary to standard Diffusion, where either of the noise or the original image is predicted, we predict both values. Upon selecting a time step \( t \), we sample a corrupted version of the image using Equation~\ref{eq:diffusionsample}. The sampled \( x_t \) is subsequently masked according to MAE protocols, resulting in \( x_t^M \). The training objective then looks like the following:

\begin{equation}
\label{Noised_Loss}
\mathcal{L}(\theta)_{t \ge 1} = \mathbb{E}_{t, \epsilon, x_0, M} \left[ \left\| M \odot (x_0 - x_{\theta}(x_t^M, t)) \right\|_2^2 + \left\| (1 - M) \odot (\epsilon - \epsilon_{\theta}(x_t^M, t)) \right\|_2^2 \right],
\end{equation}

where \( M \) refers to the random mask applied to noised images \( x_t \), \( \epsilon_{\theta} \) indicates the encoder-decoder prediction of the Diffusion noise \( \epsilon \) and \( x_{{\theta}} \) is the \( x_0 \) prediction. Because it is generally very difficult to predict the noise added to masked tokens, the loss for the \( \epsilon \) prediction is computed only on visible tokens while the \( x_0 \) prediction loss is taken on masked tokens. 

In addition to predicting both the noise and the original images, we modify the variance schedule of the diffusion process to better accommodate the representational needs. In the original objective, training on noisy images at all steps can hinder the recovery of representations during inference as the model always expects a noisy input to produce a representation. To avoid this problem, we include an additional `no noise' reconstruction step in diffusion, effectively $\beta_0 = 0.0$, in the training process, which adds no noise to the images but still masks randomly. Consequently, our adjusted variance schedule is defined as \(\beta_0, \ldots, \beta_T\). The loss for the no noise reconstruction step mimics the MAE objective exactly, denoted as follows:

\begin{equation}
\label{MAE_Loss}
\mathcal{L}(\theta)_{t=0} = \mathbb{E}_{x_0, M_0} \left[ \norm{ M_{0} \odot (x_0 - x_{\theta}(x_0^{M_0}, 0)) }_2^2 \right].
\end{equation}

Since no-noise samples pose easier challenges with respect to reconstruction as compared to noisy samples, we implement varying masking ratios: a higher ratio for no-noise samples (\( m_{t=0} \)) compared to noised samples (\( m_{t \ge 1} \) ). To leverage this higher masking ratio and to improve representation performance, we allocate a significantly higher sampling ratio (denoted \( r_{t=0} \)) for this reconstruction step at $t=0$, contrary to what is typically done in diffusion where the time steps are uniformly sampled. The overall objective for UMD looks like the following:

\begin{equation} \label{UMD_Loss}
\mathcal{L}(\theta)_{\text{UMD}} = r_{t=0} \cdot \mathcal{L}(\theta)_{t=0} +  (1 - r_{t=0}) \cdot \mathcal{L}(\theta)_{t \ge 1}.
\end{equation}

This objective unifies the two successful auto-encoder methods within one framework which jointly learns features useful for representation tasks as well as generation tasks. UMD allows for seamless play between DiT and MAE. When \( r_{t=0} = 1.0\) and \( m_{t=0} = 0.75\), we recover MAE~\citep{he2022masked}. When \( r_{t=0} = 0.0\) and \( m_{t \ge 1} = 0.5\), we recover MaskDiT~\citep{zheng2023fast}. Finally, when \( r_{t=0} = 0.0\) and \( m_{t \ge 1} = 0.0\), we recover DiT~\citep{peebles2023scalable}. Through tuning of \( r_{t=0}\)  and \( m_{t \ge 1} \), UMD allows for learning strong features for many downstream tasks, from linear probing to class conditional generation. 

\section{Experimental Analysis}
In this section, we evaluate the Unified Masked Diffusion (UMD) method, comparing its performance in representation learning and generative tasks against established methods such as Masked Auto-Encoders (MAE) and Diffusion Transformers (DiT). Our aim is to demonstrate UMD's efficiency and effectiveness as a viable alternative to these methods. 

To assess the quality of representations learned by UMD, we adopt the linear probing methodology, a standard way to benchmark self-supervised representations \citep{chen2020simple, chen2021exploring, he2022masked, chen2024deconstructing}. This approach allows us to directly evaluate the information content of the latent representations produced by our model. Furthermore, we also test UMD’s transfer learning capabilities across OOD datasets in a few-shot linear probing test. In our generative analysis, we focus on class-conditional image generation after minimal finetuning, using the Fréchet Inception Distance (FID)~\citep{heusel2017gans} and Inception Score (IS)~\citep{salimans2016improved} to measure image quality. We also highlight the efficiency gains of UMD over DiT and MaskDiT, particularly in terms of reduced computational overhead due to its masking mechanism. 

\paragraph{Experimental Setup.} We conducted our experiments on the ImageNet-1K dataset~\citep{russakovsky2015imagenet} downscaled to 64 $\times$ 64 $\times$ 3 for pixel-level experiments and comparisons with baselines. For the diffusion beta schedule, we utilize the cosine schedule proposed by~\cite{nichol2021improved} and use 1000 beta steps and 125 sampling steps at inference for per DDIM (Sec.~\ref{prelim}) unless otherwise stated. For UMD's asymmetric encoder-decoder architecture, we use a ViT-B/4 for the encoder and shallower decoder with 4 layers, incorporating AdaLN-zero conditioning as proposed by DiT~\citep{peebles2023scalable}. We apply light random cropping (scaling between 0.8 and 1.0) and random horizontal flips. We train for 800 epochs with AdamW~\cite{kingma2014adam, loshchilov2017decoupled} optimizer with a batch size of 1024 and weight decay of 0.05, following a cosine decay learning rate schedule~\citep{loshchilov2016sgdr}. This schedule has a base rate of $\frac{1.5 \cdot \text{batch size}}{256} \times 10^{-4}$ and a warmup of 40 epochs. To produce a UMD representation, the input image \(x_0\) is fed into the encoder, and the CLS token is extracted. For other models where \(r_{t=0} = 0.0\), the image is noised at \(t=50\) (i.e., \(x_{50} = q(x_{50} | x_0)\)) following the analysis of \citep{yang2023diffusion, mukhopadhyay2023diffusion} before passing into the encoder.

\paragraph{Baselines.}
We re-implement baseline methods in our codebase and test them against UMD. We focus our comparison on these methods to avoid any confounding factors or potential re-implementation issues. Unless otherwise stated, each method uses the same training protocol and architecture described in the prior paragraph. Since specific values of $r_{t=0}$ and $m_{t \ge 1}$ lead to different instances of MAE, MaskDiT and DiT, we use these three methods as natural baselines. We re-run MAE, MaskDiT, and DiT with the hyperparameters described in Section~\ref{acmethod}. For MAE, we utilize a batch size of 4096 and do not use AdaLN conditioning to be faithful to the original implementation. We found that higher batch sizes + higher scaled LR in combination with AdaLN led to instabilities for the other methods. For UMD, we set \( r_{t=0} = 0.5\), \( m_{t \ge 1} = 0.375\), and \( m_{t=0} = 0.75\). We ablate on more hyperparemeter settings in Section~\ref{ref:Ablations}. 

\label{ref:LinearProbeResults}
\subsection{Image Classification}
We test UMD's representational performance through ImageNet classification, in both linear probing as well as few-shot transfer learning settings.

\begin{table}[ht]
\caption{\textbf{64 $\times$ 64 Imagenet-1k Results.} We evaluate our UMD method and other relevant baselines on linear probing performance and subsequently evaluate FID and IS metrics after fine-tuning on labels. UMD exhibits competitive performance with MAE in linear probing while requiring similar computational demands. UMD achieves competitive FID and IS scores when paired with CFG for fine-tuning compared with DiT. FID/IS denotes naïve class conditioning versions whereas FID/IS-G denotes CFG versions, where we use $s = 1.5$. $(\uparrow) / (\downarrow)$ denotes if higher / lower is better respectively. } \label{tab:linear_probing_results}
\vspace{2mm}
\centering
\begin{tabular}{@{}lcccccc@{}}
\toprule
Method  & v4-8 TPU Hrs. $(\downarrow)$ & 100-Shot Acc. $(\uparrow)$ & FID $(\downarrow)$ & FID-G $(\downarrow)$ & IS $(\uparrow)$ & IS-G $(\uparrow)$ \\
\midrule
MAE  & \textbf{45} & \textbf{36.6\%} & 34.1 & 26.8 & 13.4 & 18.5 \\
DiT  & 105 & 25.7\% & \textbf{21.2 }& \textbf{18.9} & \textbf{23.0} & 46.9\\ 
MaskDiT  & 72 & 26.3\% & 22.1 & 19.0 & 22.1 & 43.4\\ 
\baseline{UMD (Ours)} & \baseline{60} & \baseline{31.8\%} & \baseline{23.2} & \baseline{19.8} & \baseline{20.2} & \baseline{\textbf{63.5}} \\ 
\bottomrule
\end{tabular}
\end{table}

\paragraph{Full Imagenet Linear Probing.}
Table~\ref{tab:linear_probing_results} shows 100-shot linear probing results for all methods. This approach differs from the full (100 epochs) linear probing training done in prior work as it is trained on significantly fewer examples. However, it is much faster (5 minutes vs. 8 hours) than the standard linear-probing. Firstly, note that DiT features result in a considerable gap in performance as compared to MAE features. This gap in performance has been shown to exist in previous works as well~\citep{chen2024deconstructing}. UMD bridges this gap and remains close in performance to MAE, surpassing both DiT and MaskDiT in classification accuracy as well as computational efficiency, as shown in the reduced amount of v4-8 TPU hours required for pre-training. However, the large gap between DiT and MAE linear probing performance remains consistent in both cases. We therefore expect UMD's linear probing performance on the original sized ImageNet dataset to be close to that of MAE as well. 

\begin{figure*}[t]
    \centering
    \includegraphics[trim={0cm 0cm 0cm 0cm}, clip, width=\linewidth]{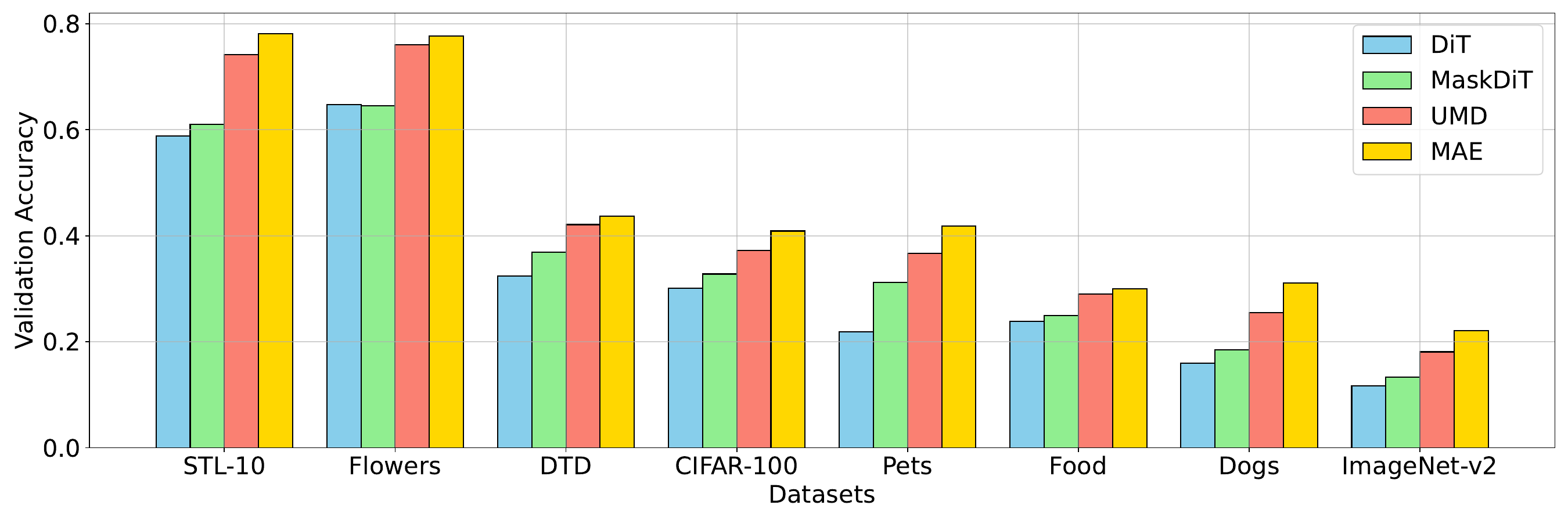}
    \caption{\textbf{Transfer learning comparison with baselines.} We run 10-shot linear probing on MAE, DiT, and UMD for the $64 \times 64 \times 3$ versions. UMD performs competitively with MAE while outperforming other diffusion methods on the different transfer datasets.}
    \label{fig:transfer_learning}
\end{figure*}

\paragraph{Transfer Learning.}
We test the effectiveness of our method in few-shot transfer learning on the following datasets: STL-10~\citep{pmlr-v15-coates11a}, Oxford Flowers~\citep{flowers102}, DTD~\citep{cimpoi14describing}, Cifar-100~\citep{krizhevsky2009learning}, Oxford Pets~\citep{parkhi12a}, food101~\citep{bossard14}, Stanford Dogs~\citep{KhoslaYaoJayadevaprakashFeiFei_FGVC2011}, and ImageNet-v2~\citep{recht2019imagenet}. We take 10 examples per class and fit a linear probe on the representations. Results are in Figure~\ref{fig:transfer_learning}. UMD remains competitive with MAE while outperforming other baselines consistently in few-shot tasks, which indicates how UMD captures a more robust representation than other diffusion models for transfer purposes.

\subsection{Image Generation}

In this section, we evaluate UMD's capability for generation tasks following a fine-tuning paradigm. Particularly, we focus on class conditional generation. Note that typically in such a setting, generative models such as DiT are trained from scratch with class labels. We, on the other hand, only introduce labels at the finetuning stage so as to keep a common pretraining paradigm for both representation and generation tasks. Arguably, our paradigm is more useful in practice since it allows for training on more easily available, label-free datasets, and then finetuning on a labelled dataset in much fewer samples. The fine-tuning process is designed to be quick and simple, as UMD already has the necessary features for generation, needing only alignment with the labels. We fine-tune the model on un-masked labeled images and sample using classifier-free guidance (CFG)~\citep{nichol2021improved, ho2022classifier} with $s=1.5$ and a 10\% dropout rate for 50 epochs. For all methods, this takes about 8 v4-8 TPU hours on our implementation. We add an EMA schedule to the parameters for use in inference time sampling as is done in DiT~\citep{peebles2023scalable}.
Performance is assessed using FID and IS metrics over 50k samples. 
The results are shown in Table~\ref{tab:linear_probing_results}. MAE generally scores the worst in FID and IS, while UMD achieves competitive scores with DiT, demonstrating UMD's strong performance in both representation and generative benchmarks.

\begin{figure*}[t]
    \centering
    \makebox[\textwidth][c]{
        \includegraphics[trim={0cm 0cm 0cm 0cm}, clip, width=1.2\linewidth]{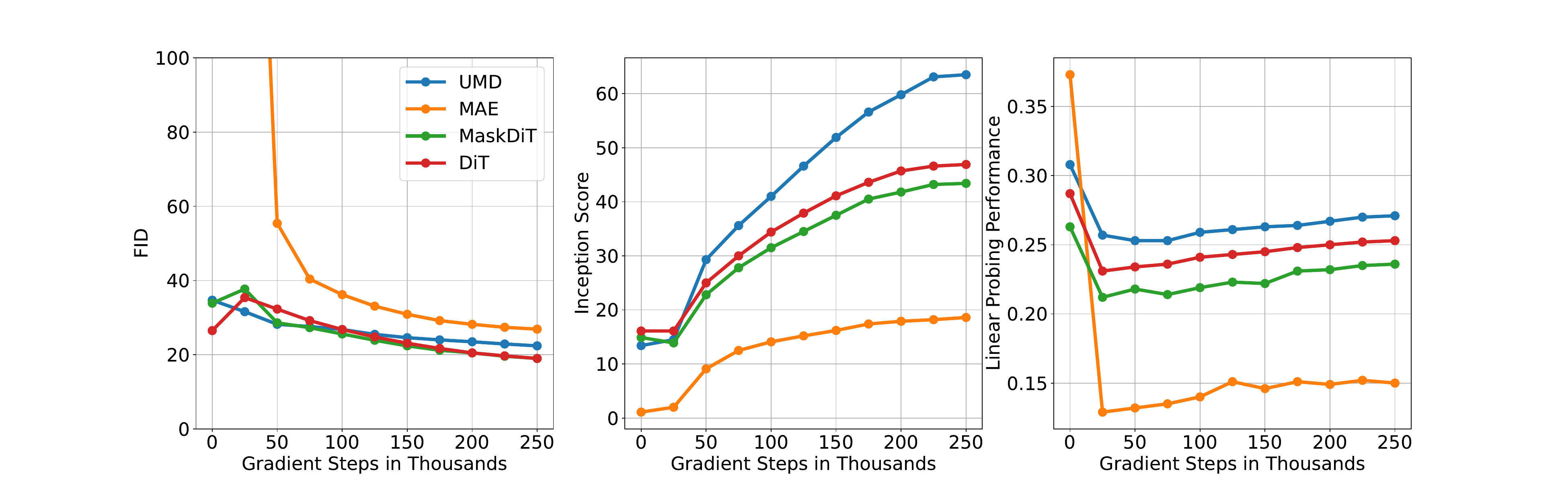}
    }
    \caption{\textbf{FID and Linear Probing Results over Fine-Tuning.} We fine-tune our baseline methods and UMD on labeled images for use in class-conditional generation. We report the 10k-FID/IS over gradient steps as well as the 100-shot linear probing performance of the representation layer. UMD remains competitive with DiT and MaskDiT in FID/IS performance and maintains its representation performance compared to MAE. }
    \label{fig:Finetune_Over_Time}
\end{figure*}

\paragraph{FID and Representation Performance over Fine-Tuning.} We compute 10k-FID/IS over fine-tuning to demonstrate the speed of FID and IS convergence. Results are in Figure~\ref{fig:Finetune_Over_Time}. Although not strictly self-supervised due to label usage, we also conduct a few-shot linear-probe to assess how representation quality changes across fine-tuning. UMD quickly converges on FID and IS, whereas MAE requires more time to improve its scores. Finally, most methods maintain their representation performance over fine-tuning, except for MAE which collapses over training. This indicates that re-aligning MAE for generations is destructive to the networks representation. 

\begin{figure*}[t]
    \centering
    \includegraphics[trim={17cm 13cm 24cm 10cm}, clip, width=0.95\linewidth]{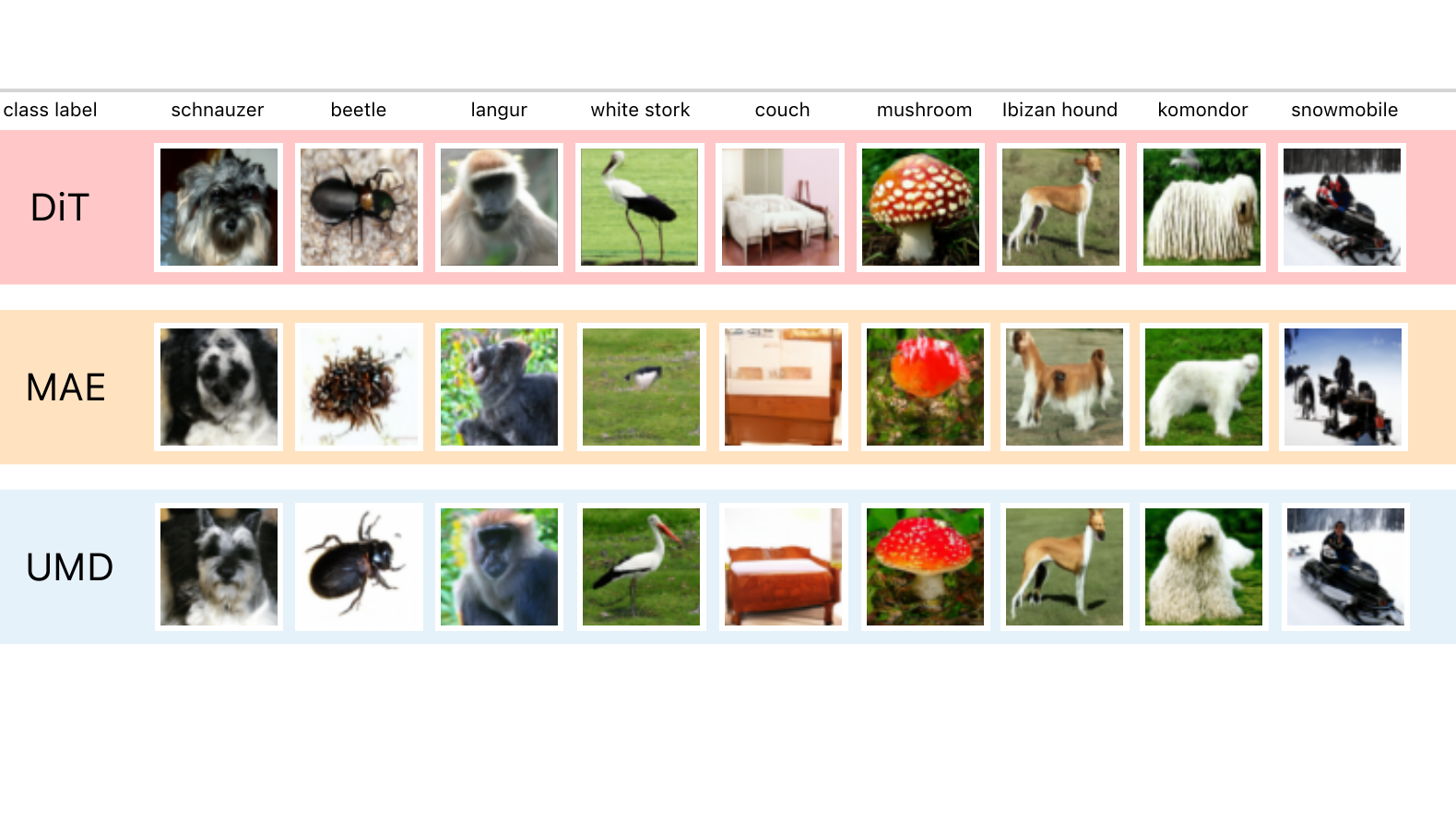}
    \vspace{-1cm}
    \caption{\textbf{Samples after finetuning.} We compare class conditioned samples generated from DiT, MAE and UMD after finetuning. Although MAE leads to a low FID score, the actual samples do not look as coherent as that of DiT and UMD, as evident by its low inception score as well.}
    \label{fig:qualitative}
\end{figure*}

\paragraph{Qualitative Analysis.} We visualize example samples from all baselines to highlight the strengths and weaknesses of each method. These samples were generated with a CFG $s=4.0$, as shown in Figure~\ref{fig:qualitative}. UMD and DiT generations are coherent and adhere well to their labels, while MAE samples are difficult to discern, despite MAE achieving a similar FID score to these methods. This indicates that MAE's features are not as effective for generation tasks. We include additional uncurated samples from UMD in Figure~\ref{fig:pixel_1}.

\subsection{Ablations}
\label{ref:Ablations}
In this section, we include additional ablations to validate the effectiveness of UMD and test sensitivity of our proposed hyperparameters: \( r_{t=0}\) and \( m_{t \ge 1}\). In particular, we include the following versions: \textbf{1}. \textit{Only Reconstruction} which uses just the reconstruction objective from Eq.~\ref{Noised_Loss}, \textbf{2}. \textit{Only Noise Prediction} which uses just the noise prediction objective from Eq.~\ref{Noised_Loss}, \textbf{3}. \textit{Reduced $t=0$  Sampling} which samples the additional $t=0$ step much less frequently, \textbf{4}. \textit{Increased Noised Masking} which uses a much higher masking ratio on the noisy samples, and \textbf{5}. \textit{No Noised Masking} which uses no masking on the noisy samples, thus effectively acting as a MAE model for $t=0$ and a DiT model for $t \ge 1$ steps. \textbf{6}. \textit{No AdaLN} which removes the AdaLN conditioning in favor of placing the condition token within the sequence, which reduces total parameters and increases training speed.

\begin{table}[ht]
\caption{\textbf{Ablation Results on 64 $\times$ 64 ImageNet-1k.} We evaluate our base UMD model and other relevant ablations.  We report the 100-shot linear probing performance on imagenet after pre-training and FID-G/IS-G ($s=1.5$) scores after fine-tuning for 50 epochs on labeled images. We report samples using the $x_0$ prediction and $\epsilon$ prediction. Our base selection of hyperparameters represents the best balance in speed and performance in downstream representation and generation tasks.} \label{tab:ablation_results}
\vspace{2mm}
\centering
\begin{tabular}{@{}lcccccccc@{}}
\toprule
Metric &  Base & $x_0$ & $\epsilon$ & Low \(r_{t=0}\) & High \(m_{t \ge 1}\) & \(m_{t \ge 1}=0.0\) & No AdaLN\\ 
\midrule
TPU Hours & 60 & 60 & 60 & 70 & 48 & 74 & 52\\
LP & 30.9\% & 31.3\% & 31.5\% & 27.7\% & 28.9\% & 32.7\% & 26.1\%\\
FID ($\epsilon$, $x_0$)  & 23.2,20.9 & -,26.7 & 25.0,- & 22.3,20.0 & 27.5,23.3 & 21.1,18.8 & 25.4,24.7\\ 
IS ($\epsilon$, $x_0$) & 63.5,62.2 & -,43.0 & 49.0,- & 64.7,63.4 & 53.2,50.5 & 65.1,63.6 & 42.5,40.8\\ 
\bottomrule
\end{tabular}
\end{table}

We report results in Table~\ref{tab:ablation_results}. We train each ablation for 800 epochs and conduct 100-shot linear probing on ImageNet-1k. Firstly, we observe that the base UMD model achieves the a strong overall score in terms of both representation performance (100-shot LP score) and generative performance (FID/IS scores). In particular, using either of the $x_0$ or $\epsilon$ reconstruction produces good representations but suffers from generative performance when finetuning. This result highlights how using both objectives (Eq.~\ref{Noised_Loss}) at the same time is critical for maintaining good generative performance when finetuning such a masked + noised model. Furthermore, using a lower sampling ratio for the first $t=0$ reconstruction step leads to the representation performance suffering. This makes intuitive sense as the first step provides the major representational benefits in our case. A higher masking ratio for $t \ge 1$ steps leads to the generative performance suffering since the diffusion model now sees much fewer samples of noisy patches and more samples of masked patches. Finally, when we switch off masking completely for $t \ge 1$ steps, we observe the best overall performance, marginally better than even the base model. However, in doing so we lose out on the computational benefits since the model needs to process all patches for every noisy sample. Since we only observe a marginal improvement, we stick to advocating for the base model as the best model among all versions, given it's superior computational benefit in processing the input. Finally, removing AdaLN conditioning worsens performance in both linear probing and generation, which indicates that AdaLN is important to the architecture of UMD. 

\subsection{Latent Diffusion}

\begin{table}[ht]
\caption{\textbf{256 $\times$ 256 Imagenet-1k Results.} We evaluate our UMD-L/2 method on linear probing and high resolution synthesis in Imagenet-1k. We compare this method to DiT trained from scratch with labels for image generation and MAE train on pixels for representation learning. UMD maintains its competitive performance to MAE and DiT in each respective domain. A $^*$ indicates that these epochs included labels during training. FID/IS denotes naïve class conditioning versions whereas FID/IS-G denotes CFG versions, where we use $s = 1.5$.} \label{tab:high_resolution_results}
\vspace{2mm}
\centering
\begin{tabular}{@{}lccccccc@{}}
\toprule
Method & Epochs & v4-TPU Hrs. & 100-Shot Acc. & FID & IS & FID-G & IS-G \\ 
\midrule
MAE & 400 & 160 & 38.8\% & - & - & - & -\\
DiT & 400$^*$ & 500 & 37.9\% & 9.60 & 101.84 & 4.31 & 274.5\\
UMD & 400 + 50$^*$ & 400 & 44.6\% & 14.5 & 78.4 & 4.86 & 201.8\\ 
MAE & 800 & 320 & 51.1\% & - & - & - & - \\
UMD & 800 + 50$^*$ & 850 & 54.4\% & 12.9 & 82.0 & 3.96 & 212.6\\
\bottomrule
\end{tabular}
\end{table}

Typically, large-scale diffusion models are trained in latent space to enable high-resolution image synthesis. We test UMD in this setting showing how it is competitive with DiT and MAE in each respective domain of representation learning and class-conditional generation. Following the approach in DiT, we use a VAE~\citep{kingma2013auto} to transform a high-resolution $256 \times 256 \times 3$ image into a $32 \times 32 \times 4$ latent space. We then apply our same training protocol as in Table~\ref{tab:linear_probing_results} except we use a ViT-L/2 as our encoder and an 8 layer decoder with the same width and number of heads (1024 and 16 respectively). Furthermore we use a linear beta schedule with a 1000 steps as in DDPM~\citep{ho2020denoising} with 250 sampling steps following the DDIM sampler. To compare against UMD and showcase the versatility of our codebase which we release with this paper, we train MAE-L/16 for 800 epochs on the full pixel $256 \times 256 \times 3$ images without latent preprocessing. We also train DiT-L/2 for 400 epochs with labels from scratch to compare for generation performance. These methods require simple hyperparameter changes to our underlying method. Results are Table~\ref{tab:high_resolution_results}.

\begin{figure*}[t]
    \centering
        \includegraphics[width=1\linewidth]{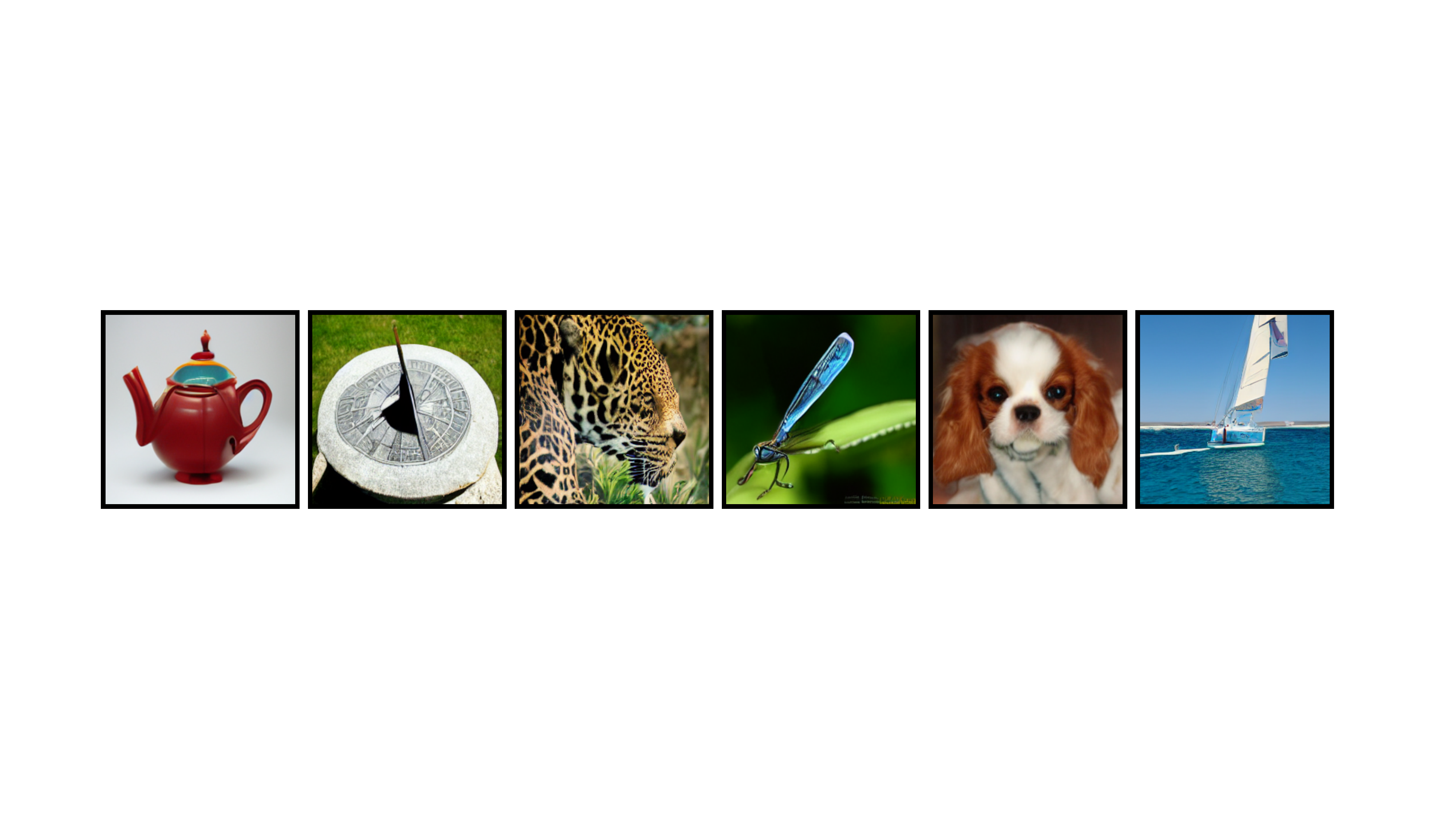}
        \vspace{-3cm}
    \caption{\textbf{Latent Unified Masked Diffusion Samples.} We pre-trained UMD on latent diffusion for
800 epochs on ImageNet and present selected samples after fine-tuning for 50 epochs. UMD achieves strong generations with a CFG scale of $s = 4.0$.}
    \label{fig:UMD_Latent_Samples}
\end{figure*}

UMD achieves better linear probing results at 400 and 800 epochs when compared to DiT and MAE. Furthermore, UMD achieves competitive FID and IS metrics compared to DiT trained with labels from scratch. This shows how UMD is an effective replacement for each method. It's important to note that UMD has seen only 50 epochs of labeled data and that the VAE plays a role in improving performance as it's seen more data ~\citep{li2023mage}. We provide select samples of fine-tuned UMD-L/2 in Figure~\ref{fig:UMD_Latent_Samples} as well as uncurrated samples of UMD-L/2 and DiT-L/2 Figure~\ref{fig:latent_1} and Figure~\ref{fig:latent_2} respectively.

\section{Limitations and Future Work}

This paper introduces a sophisticated way to incorporate benefits from two very similar looking methods in MAE and DiT, so as to maintain both representational and generative benefits. UMD however still performs image corruption either through fine-grained noise addition (like in Diffusion) or through much coarser patch masking (like in MAE), which can seem to be a very ad-hoc way of handling this issue. Having observed the benefits that UMD provides, it would be nice to merge these two corruption schemes into a very general `gating' mechanism~\citep{tomar2024ignorance}, one that smoothly goes from fine to coarser noise addition, with the special case of zero noise acting as the masking operation. In addition, it would be very interesting if such a corruption schedule can be learnt alongside the UMD model, instead of keeping it fixed \textit{a priori}. Also, our latent UMD implementation did not achieve the same speed ups as in the pixel variation over DiT. We suspect that the VAE prepossessing presents a significant overhead.
\vspace{-0.2cm}
\section{Conclusion}
This paper introduced Unified Masked Diffusion (UMD), a method equipped with both representational and generative capabilities. UMD works by following an augmented diffusion schedule where an additional, masking-based reconstruction step is included within an asymmetric encoder-decoder Diffusion model. Masking is further added to the noisy steps of the Diffusion schedule so as to provide computational benefits while pretraining. UMD remains closely matched with MAE with respect to linear probing representation performance, while also matching DiT in terms of generative performance when fine-tuning on labels, all while having a faster training time. 

\section{Acknowledgments}
This research was supported by the TRC program from Google Cloud Computing (GCP) as well as Akash Network (\texttt{https://akash.network/}). 

\bibliographystyle{plainnat}
\bibliography{bibliography.bib}

\begin{thebibliography}{63}
\providecommand{\natexlab}[1]{#1}
\providecommand{\url}[1]{\texttt{#1}}
\expandafter\ifx\csname urlstyle\endcsname\relax
  \providecommand{\doi}[1]{doi: #1}\else
  \providecommand{\doi}{doi: \begingroup \urlstyle{rm}\Url}\fi

\bibitem[Abstreiter et~al.(2021)Abstreiter, Mittal, Bauer, Sch{\"o}lkopf, and Mehrjou]{abstreiter2021diffusion}
Korbinian Abstreiter, Sarthak Mittal, Stefan Bauer, Bernhard Sch{\"o}lkopf, and Arash Mehrjou.
\newblock Diffusion-based representation learning.
\newblock \emph{arXiv preprint arXiv:2105.14257}, 2021.

\bibitem[Ba et~al.(2016)Ba, Kiros, and Hinton]{ba2016layer}
Jimmy~Lei Ba, Jamie~Ryan Kiros, and Geoffrey~E Hinton.
\newblock Layer normalization.
\newblock \emph{arXiv preprint arXiv:1607.06450}, 2016.

\bibitem[Bai et~al.(2023)Bai, Geng, Mangalam, Bar, Yuille, Darrell, Malik, and Efros]{bai2023sequential}
Yutong Bai, Xinyang Geng, Karttikeya Mangalam, Amir Bar, Alan Yuille, Trevor Darrell, Jitendra Malik, and Alexei~A Efros.
\newblock Sequential modeling enables scalable learning for large vision models.
\newblock \emph{arXiv preprint arXiv:2312.00785}, 2023.

\bibitem[Bao et~al.(2021)Bao, Dong, Piao, and Wei]{bao2021beit}
Hangbo Bao, Li~Dong, Songhao Piao, and Furu Wei.
\newblock Beit: Bert pre-training of image transformers.
\newblock \emph{arXiv preprint arXiv:2106.08254}, 2021.

\bibitem[Beyer et~al.(2022)Beyer, Zhai, and Kolesnikov]{big_vision}
Lucas Beyer, Xiaohua Zhai, and Alexander Kolesnikov.
\newblock Big vision.
\newblock \url{https://github.com/google-research/big_vision}, 2022.

\bibitem[Bossard et~al.(2014)Bossard, Guillaumin, and Van~Gool]{bossard14}
Lukas Bossard, Matthieu Guillaumin, and Luc Van~Gool.
\newblock Food-101 -- mining discriminative components with random forests.
\newblock In \emph{European Conference on Computer Vision}, 2014.

\bibitem[Bradbury et~al.(2018)Bradbury, Frostig, Hawkins, Johnson, Leary, Maclaurin, Necula, Paszke, Vander{P}las, Wanderman-{M}ilne, and Zhang]{jax2018github}
James Bradbury, Roy Frostig, Peter Hawkins, Matthew~James Johnson, Chris Leary, Dougal Maclaurin, George Necula, Adam Paszke, Jake Vander{P}las, Skye Wanderman-{M}ilne, and Qiao Zhang.
\newblock {JAX}: composable transformations of {P}ython+{N}um{P}y programs, 2018.
\newblock URL \url{http://github.com/google/jax}.

\bibitem[Caron et~al.(2021)Caron, Touvron, Misra, J{\'e}gou, Mairal, Bojanowski, and Joulin]{caron2021emerging}
Mathilde Caron, Hugo Touvron, Ishan Misra, Herv{\'e} J{\'e}gou, Julien Mairal, Piotr Bojanowski, and Armand Joulin.
\newblock Emerging properties in self-supervised vision transformers.
\newblock In \emph{Proceedings of the IEEE/CVF international conference on computer vision}, pages 9650--9660, 2021.

\bibitem[Chang et~al.(2022)Chang, Zhang, Jiang, Liu, and Freeman]{chang2022maskgit}
Huiwen Chang, Han Zhang, Lu~Jiang, Ce~Liu, and William~T Freeman.
\newblock Maskgit: Masked generative image transformer.
\newblock In \emph{Proceedings of the IEEE/CVF Conference on Computer Vision and Pattern Recognition}, pages 11315--11325, 2022.

\bibitem[Chen et~al.(2020)Chen, Kornblith, Norouzi, and Hinton]{chen2020simple}
Ting Chen, Simon Kornblith, Mohammad Norouzi, and Geoffrey Hinton.
\newblock A simple framework for contrastive learning of visual representations.
\newblock In \emph{International conference on machine learning}, pages 1597--1607. PMLR, 2020.

\bibitem[Chen and He(2021)]{chen2021exploring}
Xinlei Chen and Kaiming He.
\newblock Exploring simple siamese representation learning.
\newblock In \emph{Proceedings of the IEEE/CVF conference on computer vision and pattern recognition}, pages 15750--15758, 2021.

\bibitem[Chen et~al.(2024)Chen, Liu, Xie, and He]{chen2024deconstructing}
Xinlei Chen, Zhuang Liu, Saining Xie, and Kaiming He.
\newblock Deconstructing denoising diffusion models for self-supervised learning.
\newblock \emph{arXiv preprint arXiv:2401.14404}, 2024.

\bibitem[Cimpoi et~al.(2014)Cimpoi, Maji, Kokkinos, Mohamed, , and Vedaldi]{cimpoi14describing}
M.~Cimpoi, S.~Maji, I.~Kokkinos, S.~Mohamed, , and A.~Vedaldi.
\newblock Describing textures in the wild.
\newblock In \emph{Proceedings of the {IEEE} Conf. on Computer Vision and Pattern Recognition ({CVPR})}, 2014.

\bibitem[Coates et~al.(2011)Coates, Ng, and Lee]{pmlr-v15-coates11a}
Adam Coates, Andrew Ng, and Honglak Lee.
\newblock An analysis of single-layer networks in unsupervised feature learning.
\newblock In Geoffrey Gordon, David Dunson, and Miroslav Dudík, editors, \emph{Proceedings of the Fourteenth International Conference on Artificial Intelligence and Statistics}, volume~15 of \emph{Proceedings of Machine Learning Research}, pages 215--223, Fort Lauderdale, FL, USA, 11--13 Apr 2011. PMLR.
\newblock URL \url{https://proceedings.mlr.press/v15/coates11a.html}.

\bibitem[Devlin et~al.(2018)Devlin, Chang, Lee, and Toutanova]{devlin2018bert}
Jacob Devlin, Ming-Wei Chang, Kenton Lee, and Kristina Toutanova.
\newblock Bert: Pre-training of deep bidirectional transformers for language understanding.
\newblock \emph{arXiv preprint arXiv:1810.04805}, 2018.

\bibitem[Dhariwal and Nichol(2021)]{dhariwal2021diffusion}
Prafulla Dhariwal and Alexander Nichol.
\newblock Diffusion models beat gans on image synthesis.
\newblock \emph{Advances in neural information processing systems}, 34:\penalty0 8780--8794, 2021.

\bibitem[Dosovitskiy et~al.(2020)Dosovitskiy, Beyer, Kolesnikov, Weissenborn, Zhai, Unterthiner, Dehghani, Minderer, Heigold, Gelly, et~al.]{dosovitskiy2020image}
Alexey Dosovitskiy, Lucas Beyer, Alexander Kolesnikov, Dirk Weissenborn, Xiaohua Zhai, Thomas Unterthiner, Mostafa Dehghani, Matthias Minderer, Georg Heigold, Sylvain Gelly, et~al.
\newblock An image is worth 16x16 words: Transformers for image recognition at scale.
\newblock \emph{arXiv preprint arXiv:2010.11929}, 2020.

\bibitem[Esser et~al.(2021)Esser, Rombach, and Ommer]{esser2021taming}
Patrick Esser, Robin Rombach, and Bjorn Ommer.
\newblock Taming transformers for high-resolution image synthesis.
\newblock In \emph{Proceedings of the IEEE/CVF conference on computer vision and pattern recognition}, pages 12873--12883, 2021.

\bibitem[Gao et~al.(2023)Gao, Zhou, Cheng, and Yan]{gao2023masked}
Shanghua Gao, Pan Zhou, Ming-Ming Cheng, and Shuicheng Yan.
\newblock Masked diffusion transformer is a strong image synthesizer.
\newblock In \emph{Proceedings of the IEEE/CVF International Conference on Computer Vision}, pages 23164--23173, 2023.

\bibitem[Geng et~al.(2022)Geng, Liu, Lee, Schuurams, Levine, and Abbeel]{geng2022multimodal}
Xinyang Geng, Hao Liu, Lisa Lee, Dale Schuurams, Sergey Levine, and Pieter Abbeel.
\newblock Multimodal masked autoencoders learn transferable representations.
\newblock \emph{arXiv preprint arXiv:2205.14204}, 2022.

\bibitem[Goodfellow et~al.(2014)Goodfellow, Pouget-Abadie, Mirza, Xu, Warde-Farley, Ozair, Courville, and Bengio]{goodfellow2014generative}
Ian Goodfellow, Jean Pouget-Abadie, Mehdi Mirza, Bing Xu, David Warde-Farley, Sherjil Ozair, Aaron Courville, and Yoshua Bengio.
\newblock Generative adversarial nets.
\newblock \emph{Advances in neural information processing systems}, 27, 2014.

\bibitem[Grill et~al.(2020)Grill, Strub, Altch{\'e}, Tallec, Richemond, Buchatskaya, Doersch, Avila~Pires, Guo, Gheshlaghi~Azar, et~al.]{grill2020bootstrap}
Jean-Bastien Grill, Florian Strub, Florent Altch{\'e}, Corentin Tallec, Pierre Richemond, Elena Buchatskaya, Carl Doersch, Bernardo Avila~Pires, Zhaohan Guo, Mohammad Gheshlaghi~Azar, et~al.
\newblock Bootstrap your own latent-a new approach to self-supervised learning.
\newblock \emph{Advances in neural information processing systems}, 33:\penalty0 21271--21284, 2020.

\bibitem[He et~al.(2016)He, Zhang, Ren, and Sun]{he2016deep}
Kaiming He, Xiangyu Zhang, Shaoqing Ren, and Jian Sun.
\newblock Deep residual learning for image recognition.
\newblock In \emph{Proceedings of the IEEE conference on computer vision and pattern recognition}, pages 770--778, 2016.

\bibitem[He et~al.(2020)He, Fan, Wu, Xie, and Girshick]{he2020momentum}
Kaiming He, Haoqi Fan, Yuxin Wu, Saining Xie, and Ross Girshick.
\newblock Momentum contrast for unsupervised visual representation learning.
\newblock In \emph{Proceedings of the IEEE/CVF conference on computer vision and pattern recognition}, pages 9729--9738, 2020.

\bibitem[He et~al.(2022)He, Chen, Xie, Li, Doll{\'a}r, and Girshick]{he2022masked}
Kaiming He, Xinlei Chen, Saining Xie, Yanghao Li, Piotr Doll{\'a}r, and Ross Girshick.
\newblock Masked autoencoders are scalable vision learners.
\newblock In \emph{Proceedings of the IEEE/CVF conference on computer vision and pattern recognition}, pages 16000--16009, 2022.

\bibitem[Heusel et~al.(2017)Heusel, Ramsauer, Unterthiner, Nessler, and Hochreiter]{heusel2017gans}
Martin Heusel, Hubert Ramsauer, Thomas Unterthiner, Bernhard Nessler, and Sepp Hochreiter.
\newblock Gans trained by a two time-scale update rule converge to a local nash equilibrium.
\newblock \emph{Advances in neural information processing systems}, 30, 2017.

\bibitem[Ho and Salimans(2022)]{ho2022classifier}
Jonathan Ho and Tim Salimans.
\newblock Classifier-free diffusion guidance.
\newblock \emph{arXiv preprint arXiv:2207.12598}, 2022.

\bibitem[Ho et~al.(2020)Ho, Jain, and Abbeel]{ho2020denoising}
Jonathan Ho, Ajay Jain, and Pieter Abbeel.
\newblock Denoising diffusion probabilistic models.
\newblock \emph{Advances in neural information processing systems}, 33:\penalty0 6840--6851, 2020.

\bibitem[Hudson et~al.(2023)Hudson, Zoran, Malinowski, Lampinen, Jaegle, McClelland, Matthey, Hill, and Lerchner]{hudson2023soda}
Drew~A Hudson, Daniel Zoran, Mateusz Malinowski, Andrew~K Lampinen, Andrew Jaegle, James~L McClelland, Loic Matthey, Felix Hill, and Alexander Lerchner.
\newblock Soda: Bottleneck diffusion models for representation learning.
\newblock \emph{arXiv preprint arXiv:2311.17901}, 2023.

\bibitem[Karras et~al.(2019)Karras, Laine, and Aila]{karras2019style}
Tero Karras, Samuli Laine, and Timo Aila.
\newblock A style-based generator architecture for generative adversarial networks.
\newblock In \emph{Proceedings of the IEEE/CVF conference on computer vision and pattern recognition}, pages 4401--4410, 2019.

\bibitem[Khosla et~al.(2011)Khosla, Jayadevaprakash, Yao, and Fei-Fei]{KhoslaYaoJayadevaprakashFeiFei_FGVC2011}
Aditya Khosla, Nityananda Jayadevaprakash, Bangpeng Yao, and Li~Fei-Fei.
\newblock Novel dataset for fine-grained image categorization.
\newblock In \emph{First Workshop on Fine-Grained Visual Categorization, IEEE Conference on Computer Vision and Pattern Recognition}, Colorado Springs, CO, June 2011.

\bibitem[Kingma and Ba(2014)]{kingma2014adam}
Diederik~P Kingma and Jimmy Ba.
\newblock Adam: A method for stochastic optimization.
\newblock \emph{arXiv preprint arXiv:1412.6980}, 2014.

\bibitem[Kingma and Welling(2013)]{kingma2013auto}
Diederik~P Kingma and Max Welling.
\newblock Auto-encoding variational bayes.
\newblock \emph{arXiv preprint arXiv:1312.6114}, 2013.

\bibitem[Krizhevsky(2009)]{krizhevsky2009learning}
Alex Krizhevsky.
\newblock Learning multiple layers of features from tiny images.
\newblock pages 32--33, 2009.
\newblock URL \url{https://www.cs.toronto.edu/~kriz/learning-features-2009-TR.pdf}.

\bibitem[Li et~al.(2023{\natexlab{a}})Li, Chang, Mishra, Zhang, Katabi, and Krishnan]{li2023mage}
Tianhong Li, Huiwen Chang, Shlok Mishra, Han Zhang, Dina Katabi, and Dilip Krishnan.
\newblock Mage: Masked generative encoder to unify representation learning and image synthesis.
\newblock In \emph{Proceedings of the IEEE/CVF Conference on Computer Vision and Pattern Recognition}, pages 2142--2152, 2023{\natexlab{a}}.

\bibitem[Li et~al.(2023{\natexlab{b}})Li, Katabi, and He]{li2023self}
Tianhong Li, Dina Katabi, and Kaiming He.
\newblock Self-conditioned image generation via generating representations.
\newblock \emph{arXiv preprint arXiv:2312.03701}, 2023{\natexlab{b}}.

\bibitem[Liu et~al.(2024)Liu, Yan, Zaharia, and Abbeel]{liu2024world}
Hao Liu, Wilson Yan, Matei Zaharia, and Pieter Abbeel.
\newblock World model on million-length video and language with ringattention.
\newblock \emph{arXiv preprint arXiv:2402.08268}, 2024.

\bibitem[Loshchilov and Hutter(2016)]{loshchilov2016sgdr}
Ilya Loshchilov and Frank Hutter.
\newblock Sgdr: Stochastic gradient descent with warm restarts.
\newblock \emph{arXiv preprint arXiv:1608.03983}, 2016.

\bibitem[Loshchilov and Hutter(2017)]{loshchilov2017decoupled}
Ilya Loshchilov and Frank Hutter.
\newblock Decoupled weight decay regularization.
\newblock \emph{arXiv preprint arXiv:1711.05101}, 2017.

\bibitem[Mukhopadhyay et~al.(2023)Mukhopadhyay, Gwilliam, Agarwal, Padmanabhan, Swaminathan, Hegde, Zhou, and Shrivastava]{mukhopadhyay2023diffusion}
Soumik Mukhopadhyay, Matthew Gwilliam, Vatsal Agarwal, Namitha Padmanabhan, Archana Swaminathan, Srinidhi Hegde, Tianyi Zhou, and Abhinav Shrivastava.
\newblock Diffusion models beat gans on image classification.
\newblock \emph{arXiv preprint arXiv:2307.08702}, 2023.

\bibitem[Nichol and Dhariwal(2021)]{nichol2021improved}
Alexander~Quinn Nichol and Prafulla Dhariwal.
\newblock Improved denoising diffusion probabilistic models.
\newblock In \emph{International conference on machine learning}, pages 8162--8171. PMLR, 2021.

\bibitem[Nilsback and Zisserman(2008)]{flowers102}
Maria-Elena Nilsback and Andrew Zisserman.
\newblock Automated flower classification over a large number of classes.
\newblock In \emph{2008 Sixth Indian Conference on Computer Vision, Graphics \& Image Processing}, pages 722--729, 2008.
\newblock \doi{10.1109/ICVGIP.2008.47}.

\bibitem[Oord et~al.(2017)Oord, Vinyals, and Kavukcuoglu]{oord2017neural}
Aaron van~den Oord, Oriol Vinyals, and Koray Kavukcuoglu.
\newblock Neural discrete representation learning.
\newblock \emph{arXiv preprint arXiv:1711.00937}, 2017.

\bibitem[Oord et~al.(2018)Oord, Li, and Vinyals]{oord2018representation}
Aaron van~den Oord, Yazhe Li, and Oriol Vinyals.
\newblock Representation learning with contrastive predictive coding.
\newblock \emph{arXiv preprint arXiv:1807.03748}, 2018.

\bibitem[Parkhi et~al.(2012)Parkhi, Vedaldi, Zisserman, and Jawahar]{parkhi12a}
O.~M. Parkhi, A.~Vedaldi, A.~Zisserman, and C.~V. Jawahar.
\newblock Cats and dogs.
\newblock In \emph{IEEE Conference on Computer Vision and Pattern Recognition}, 2012.

\bibitem[Peebles and Xie(2023)]{peebles2023scalable}
William Peebles and Saining Xie.
\newblock Scalable diffusion models with transformers.
\newblock In \emph{Proceedings of the IEEE/CVF International Conference on Computer Vision}, pages 4195--4205, 2023.

\bibitem[Perez et~al.(2018)Perez, Strub, De~Vries, Dumoulin, and Courville]{perez2018film}
Ethan Perez, Florian Strub, Harm De~Vries, Vincent Dumoulin, and Aaron Courville.
\newblock Film: Visual reasoning with a general conditioning layer.
\newblock In \emph{Proceedings of the AAAI conference on artificial intelligence}, volume~32, 2018.

\bibitem[Preechakul et~al.(2022)Preechakul, Chatthee, Wizadwongsa, and Suwajanakorn]{preechakul2022diffusion}
Konpat Preechakul, Nattanat Chatthee, Suttisak Wizadwongsa, and Supasorn Suwajanakorn.
\newblock Diffusion autoencoders: Toward a meaningful and decodable representation.
\newblock In \emph{Proceedings of the IEEE/CVF Conference on Computer Vision and Pattern Recognition}, pages 10619--10629, 2022.

\bibitem[Ramesh et~al.(2021)Ramesh, Pavlov, Goh, Gray, Voss, Radford, Chen, and Sutskever]{ramesh2021zero}
Aditya Ramesh, Mikhail Pavlov, Gabriel Goh, Scott Gray, Chelsea Voss, Alec Radford, Mark Chen, and Ilya Sutskever.
\newblock Zero-shot text-to-image generation.
\newblock In \emph{International Conference on Machine Learning}, pages 8821--8831. PMLR, 2021.

\bibitem[Recht et~al.(2019)Recht, Roelofs, Schmidt, and Shankar]{recht2019imagenet}
Benjamin Recht, Rebecca Roelofs, Ludwig Schmidt, and Vaishaal Shankar.
\newblock Do imagenet classifiers generalize to imagenet?
\newblock In \emph{International conference on machine learning}, pages 5389--5400. PMLR, 2019.

\bibitem[Rombach et~al.(2022)Rombach, Blattmann, Lorenz, Esser, and Ommer]{rombach2022high}
Robin Rombach, Andreas Blattmann, Dominik Lorenz, Patrick Esser, and Bj{\"o}rn Ommer.
\newblock High-resolution image synthesis with latent diffusion models.
\newblock In \emph{Proceedings of the IEEE/CVF conference on computer vision and pattern recognition}, pages 10684--10695, 2022.

\bibitem[Ronneberger et~al.(2015)Ronneberger, Fischer, and Brox]{ronneberger2015u}
Olaf Ronneberger, Philipp Fischer, and Thomas Brox.
\newblock U-net: Convolutional networks for biomedical image segmentation.
\newblock In \emph{Medical Image Computing and Computer-Assisted Intervention--MICCAI 2015: 18th International Conference, Munich, Germany, October 5-9, 2015, Proceedings, Part III 18}, pages 234--241. Springer, 2015.

\bibitem[Russakovsky et~al.(2015)Russakovsky, Deng, Su, Krause, Satheesh, Ma, Huang, Karpathy, Khosla, Bernstein, et~al.]{russakovsky2015imagenet}
Olga Russakovsky, Jia Deng, Hao Su, Jonathan Krause, Sanjeev Satheesh, Sean Ma, Zhiheng Huang, Andrej Karpathy, Aditya Khosla, Michael Bernstein, et~al.
\newblock Imagenet large scale visual recognition challenge.
\newblock \emph{International journal of computer vision}, 115:\penalty0 211--252, 2015.

\bibitem[Saharia et~al.(2022)Saharia, Chan, Saxena, Li, Whang, Denton, Ghasemipour, Gontijo~Lopes, Karagol~Ayan, Salimans, et~al.]{saharia2022photorealistic}
Chitwan Saharia, William Chan, Saurabh Saxena, Lala Li, Jay Whang, Emily~L Denton, Kamyar Ghasemipour, Raphael Gontijo~Lopes, Burcu Karagol~Ayan, Tim Salimans, et~al.
\newblock Photorealistic text-to-image diffusion models with deep language understanding.
\newblock \emph{Advances in neural information processing systems}, 35:\penalty0 36479--36494, 2022.

\bibitem[Salimans et~al.(2016)Salimans, Goodfellow, Zaremba, Cheung, Radford, and Chen]{salimans2016improved}
Tim Salimans, Ian Goodfellow, Wojciech Zaremba, Vicki Cheung, Alec Radford, and Xi~Chen.
\newblock Improved techniques for training gans.
\newblock \emph{Advances in neural information processing systems}, 29, 2016.

\bibitem[Sohl-Dickstein et~al.(2015)Sohl-Dickstein, Weiss, Maheswaranathan, and Ganguli]{sohl2015deep}
Jascha Sohl-Dickstein, Eric Weiss, Niru Maheswaranathan, and Surya Ganguli.
\newblock Deep unsupervised learning using nonequilibrium thermodynamics.
\newblock In \emph{International conference on machine learning}, pages 2256--2265. PMLR, 2015.

\bibitem[Song and Ermon(2019)]{song2019generative}
Yang Song and Stefano Ermon.
\newblock Generative modeling by estimating gradients of the data distribution.
\newblock \emph{Advances in neural information processing systems}, 32, 2019.

\bibitem[Song et~al.(2020)Song, Sohl-Dickstein, Kingma, Kumar, Ermon, and Poole]{song2020score}
Yang Song, Jascha Sohl-Dickstein, Diederik~P Kingma, Abhishek Kumar, Stefano Ermon, and Ben Poole.
\newblock Score-based generative modeling through stochastic differential equations.
\newblock \emph{arXiv preprint arXiv:2011.13456}, 2020.

\bibitem[Tomar et~al.(2024)Tomar, Islam, Taylor, Levine, and Bachman]{tomar2024ignorance}
Manan Tomar, Riashat Islam, Matthew Taylor, Sergey Levine, and Philip Bachman.
\newblock Ignorance is bliss: Robust control via information gating.
\newblock \emph{Advances in Neural Information Processing Systems}, 36, 2024.

\bibitem[Vaswani et~al.(2017)Vaswani, Shazeer, Parmar, Uszkoreit, Jones, Gomez, Kaiser, and Polosukhin]{vaswani2017attention}
Ashish Vaswani, Noam Shazeer, Niki Parmar, Jakob Uszkoreit, Llion Jones, Aidan~N Gomez, {\L}ukasz Kaiser, and Illia Polosukhin.
\newblock Attention is all you need.
\newblock \emph{Advances in neural information processing systems}, 30, 2017.

\bibitem[Wei et~al.(2023)Wei, Mangalam, Huang, Li, Fan, Xu, Wang, Xie, Yuille, and Feichtenhofer]{wei2023diffusion}
Chen Wei, Karttikeya Mangalam, Po-Yao Huang, Yanghao Li, Haoqi Fan, Hu~Xu, Huiyu Wang, Cihang Xie, Alan Yuille, and Christoph Feichtenhofer.
\newblock Diffusion models as masked autoencoders.
\newblock \emph{arXiv preprint arXiv:2304.03283}, 2023.

\bibitem[Yang and Wang(2023)]{yang2023diffusion}
Xingyi Yang and Xinchao Wang.
\newblock Diffusion model as representation learner.
\newblock In \emph{Proceedings of the IEEE/CVF International Conference on Computer Vision}, pages 18938--18949, 2023.

\bibitem[Zheng et~al.(2023)Zheng, Nie, Vahdat, and Anandkumar]{zheng2023fast}
Hongkai Zheng, Weili Nie, Arash Vahdat, and Anima Anandkumar.
\newblock Fast training of diffusion models with masked transformers.
\newblock \emph{arXiv preprint arXiv:2306.09305}, 2023.

\end{thebibliography}


\appendix

\section{Implementation Details}
We include additional details about how we trained UMD in the pixel and latent diffusion case. Our implementation is based off the big vision codebase~\citep{big_vision} which is written in Jax~\citep{jax2018github}. The MAE implementation was largely inspired by the implementation in M3AE~\citep{geng2022multimodal}. We had access to 80 $\times$ A100 GPUs via a university cluster and Akash. We also had access to 64 $\times$ v4-TPUs via the TRC program from Google Cloud. These were used over the course of 6 months for this project.

\paragraph{Architecture details.} Following DiT, we embed input discrete time steps using a 256-dimensional frequency embedding~\citep{dhariwal2021diffusion} followed with a two layer MLP trunk with SilU activations, where the hidden dimension is equal to the embedding dimension of the transformer. The transformer architecture follows the Vision Transformer (ViT)~\citep{dosovitskiy2020image} architecture. We patchify the image or latent space and use a linear convolution to produce an embedding. Then, it is passed through an encoder-decoder architecture similar to MAE~\citep{he2022masked}. This makes it so that a portion of the sequence is dropped when passed into the encoder and replaced with a masked token embedding in the decoder. A single CLS token is added at the beginning of this sequence. This architecture uses Transformer blocks~\citep{vaswani2017attention} that consist of multi-head self-attention and an MLP block that utilizes residual connections~\citep{he2016deep} and layer normalization~\citep{ba2016layer}. In each block, conditioning is passed into the adaLN layer which goes through a SiLU nonlinearity and then a linear layer that gives us the scale, shift, and mask values ($\times 6$). This modulates the sequence within the Transformer block. We use two learnable position embeddings at the encoder and decoder. The final layer, right before the convolution transpose, uses only scale and shift. To convert to an image, we use a convolution transpose. The features from the encoders CLS token are used for representation learning, while both the encoder and decoder are used for generations later.

\paragraph{Comments.} In MAE a larger batch size of 4096 coupled with a larger LR rate is used. We found in our setting that a batch size of 2048 or greater lead to instabilities during pre-training in UMD and MaskDiT. We suspect the issue is the use of AdaLN modulation. 

We also found it hard to achieve a large speed up over DiT on the latent diffusion implementation. UMD-L/2 was about 15\% faster than DiT, but this does not represent the 42\% speed up seen in the small scale $64 \times 64$ results. We suspect that the need to encode the high resolution $256 \times 256$ with a VAE slows down the overall implementation. We leave this for future work. 

\begin{center}
\begin{table}[ht]
\caption{\textbf{Pre-training Hyperparameters for UMD.}}\label{tab:hyperparam_umd_pixel}
\def\arraystretch{1.35}
\begin{tabular}{|l|c|} 
\hline
\textbf{Model Variant} & ViT-B/4 for Pixels $64 \times 64 \times 3$, ViT-L/2 for Latents $32 \times 32 \times 4$ \\
\textbf{Optimizer} & AdamW~\citep{loshchilov2017decoupled} \\
\textbf{LR}  & 6e-4 \\
\textbf{Batch Size}  & 1024 \\
\textbf{Weight Decay}  & 0.05\\
\textbf{Optimizer Momentum} & $\beta_1, \beta_2$ = 0.9, 0.95\\
\textbf{Learning Rate Schedule}  & Cosine Decay~\citep{loshchilov2016sgdr}\\
\textbf{Warmup Epochs} & 40 \\
\textbf{Training Epochs}  & 800 \\
\textbf{Gradient Clip} & 1.0 \\
$r_{t=0}$  &  0.5\\
$m_{t=0}$ & 0.75 \\
$m_{t \ge 1}$ & 0.375 \\
\textbf{Beta schedule} & \text{Cosine \citep{nichol2021improved}} for Pixels and Linear~\citep{ho2020denoising} for Latents. \\
\textbf{Number Beta Steps} & 1000 \\
\textbf{Augmentation} & RandomResizedCrop(0.8, 1.0) + Horizontal Flips \\
\hline
\end{tabular}
\end{table}
\end{center}

\paragraph{Pre-training.} For pre-training our hyperparameters are in Table~\ref{tab:hyperparam_umd_pixel}. UMD largely ports the hyperparameters used in MAE and DiT with some minor changes in batch size and LR to improve stability in training.

\begin{center}
\begin{table}[ht]
\caption{\textbf{Fine-tuning Hyperparameters for UMD.}}\label{tab:hyperparam_umd_fine}
\def\arraystretch{1.35}
\begin{tabular}{|l|c|} 
\hline
\textbf{Model Variant} & ViT-B/4 for Pixels, ViT-L/2 for Latents \\
\textbf{Optimizer} & AdamW~\citep{loshchilov2017decoupled} \\
\textbf{LR}  & 1.5e-4 \\
\textbf{Batch Size}  & 256 \\
\textbf{Weight Decay}  & 0.0\\
\textbf{Optimizer Momentum} & $\beta_1, \beta_2$ = 0.9, 0.999\\
\textbf{Learning Rate Schedule}  & Cosine Decay~\citep{loshchilov2016sgdr}\\
\textbf{Warmup Epochs} & 2.5 \\
\textbf{Training Epochs}  & 50 \\
\textbf{Gradient Clip} & 1.0 \\
$r_{t=0}$  &  0.05\\
$m_{t=0}$ & 0.75 \\
$m_{t \ge 1}$ & 0.0 \\
\textbf{Beta schedule} & \text{Cosine \citep{nichol2021improved}} for pixels and Linear~\citep{ho2020denoising} for latents. \\
\textbf{Number Beta Steps} & 1000 \\
\textbf{Number Sampling Steps} & 250 \\
\textbf{DDIM~\citep{song2020score} $\eta$} & 1.0 \\
\textbf{Augmentation} & RandomResizedCrop(0.95, 1.0) + Horizontal Flips \\
\textbf{EMA Decay Factor (Polyak)} & 0.00025\\
\hline
\end{tabular}
\end{table}
\end{center}
\paragraph{Fine-tuning.} After loading the weights from the pre-training process, we use an embedding layer to convert the discrete class tokens into an embedding the dimension of the hidden layer. These networks are added to the entire architecture. We then add the condition embedding to the time conditioned embedding before it is passed into the adaLN modulation layer. The hyperparameters for generation fine-tuning are in Table~\ref{tab:hyperparam_umd_fine}. 

\paragraph{Linear Probing Details.} For the 100-shot LP, we utilize the few-shot probe from the Big Vision~\citep{big_vision} repo, which computes the optimal linear weights based on the training data.

\paragraph{Tokenizer in Latent Diffusion.} We utilize the VQGAN~\citep{esser2021taming} tokenizer. It is a CNN based tokenizer that takes a $256 \times 256$ input and brings it down to $32 \times 32 \times 4$ embedding space. The encoder consists of 5 blocks and each block has 2 residual blocks. The quantizer then quantizes each pixel of the encoder to a 1024 codebook. The detokenizer uses 5 blocks with 2 residual blocks each. The features are upsampled by 2 per block.
\section{Example Generations}
\begin{figure*}[t]
    \centering
    \includegraphics[trim={0cm 0cm 0cm 0cm}, clip, width=\linewidth]{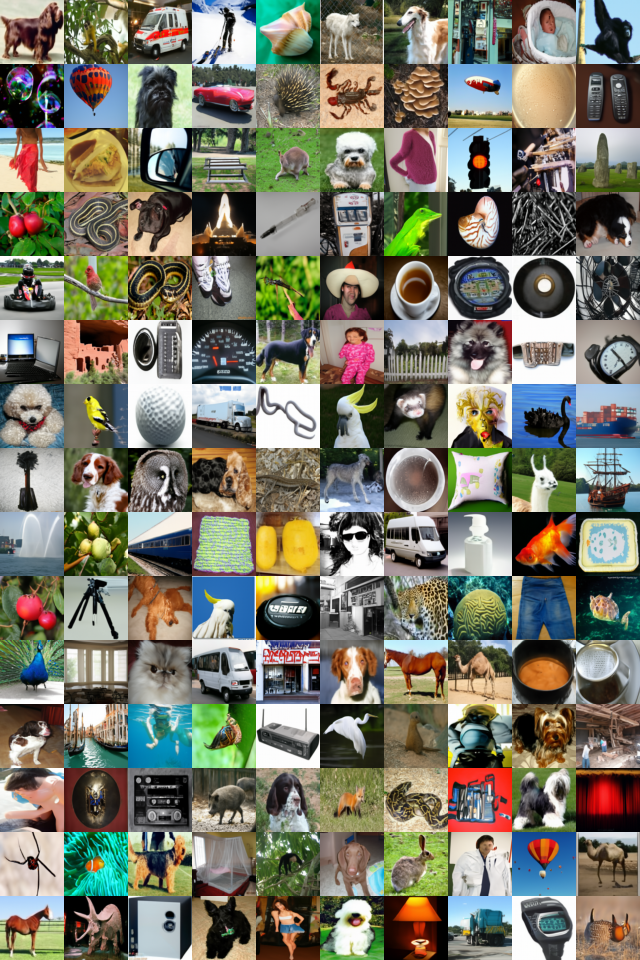}
    \caption{\textbf{Uncurated Class-Conditional Samples for $64 \times 64$ ImageNet-1k.} Randomly selected samples from the UMD-B/4 pixel diffusion model after fine-tuning.}
    \label{fig:pixel_1}
\end{figure*}

\begin{figure*}[t]
    \centering
    \includegraphics[trim={0cm 0cm 0cm 0cm}, clip, width=\linewidth]{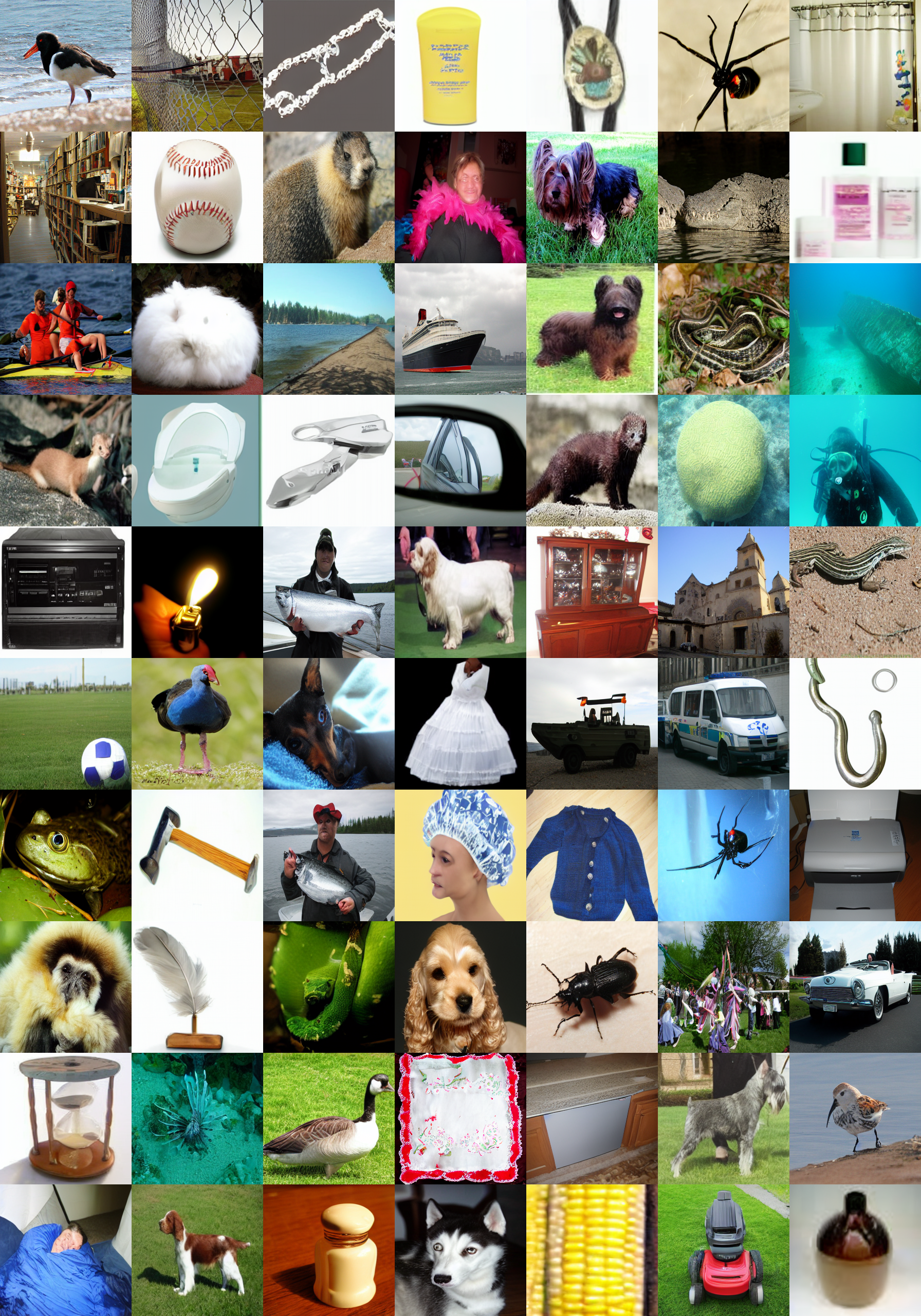}
    \caption{\textbf{Uncurated Class-Conditional Samples on $256 \times 256$ ImageNet-1k.} Randomly selected samples from the UMD-L/2 latent diffusion model after fine-tuning.}
    \label{fig:latent_1}
\end{figure*}

\begin{figure*}[t]
    \centering
    \includegraphics[trim={0cm 0cm 0cm 0cm}, clip, width=\linewidth]{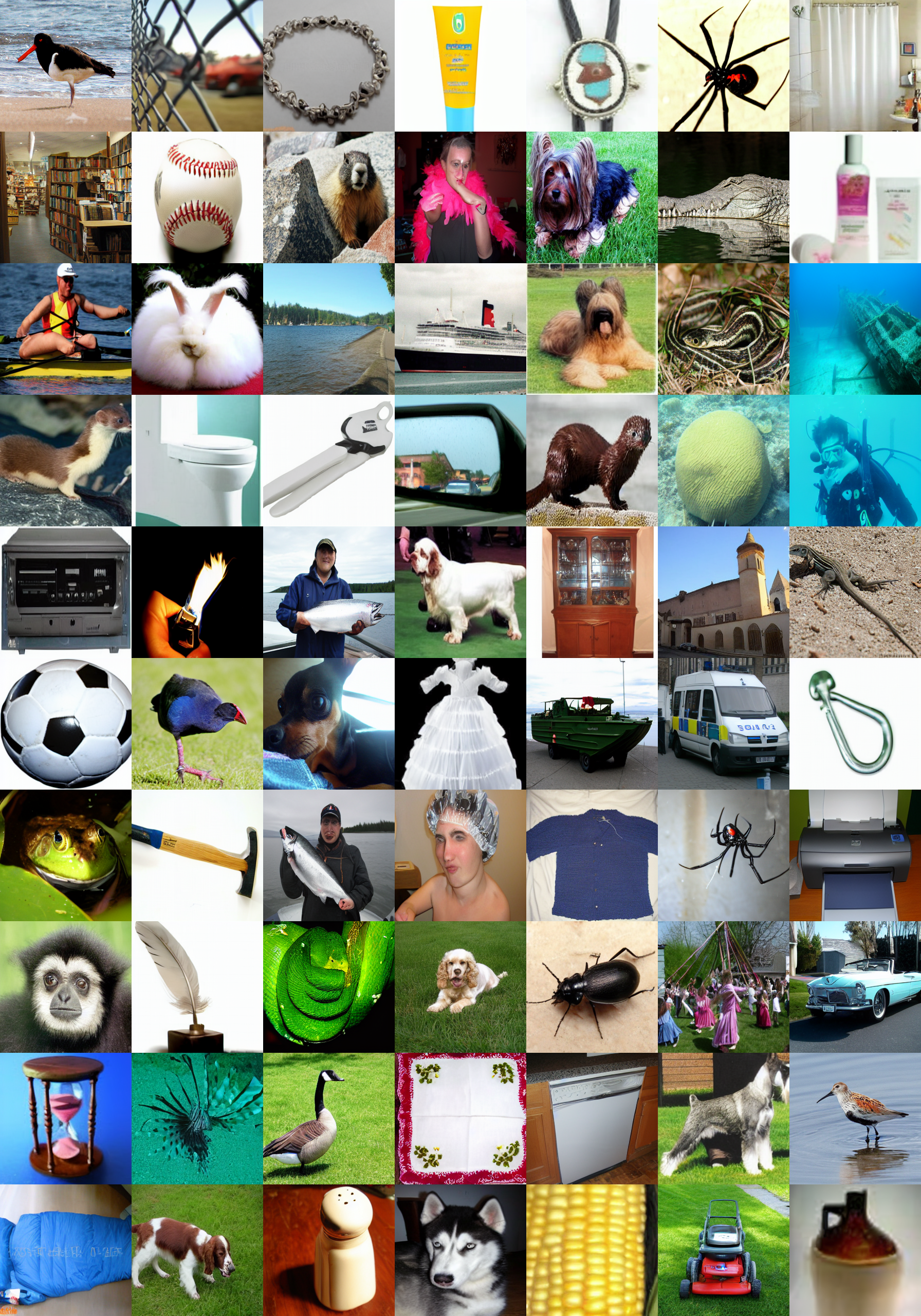}
    \caption{\textbf{Uncurated Class-Conditional Samples on $256 \times 256$ ImageNet-1k.} Randomly selected samples from the DiT-L/2 latent diffusion model after training from scratch with labels.}
    \label{fig:latent_2}
\end{figure*}

In this section we provide extra generations from UMD in both the $64 \times 64$ setting and in the $256 \times 256$ setting. These examples are randomly selected and no cherrypicking takes place. We use a CFG scale of 4.0 and the DDIM sampler with 250 steps for all examples. Figure~\ref{fig:pixel_1} and Figure~\ref{fig:latent_1} shows our examples on pixels and latents respectively. We also show DiT-L/2 samples after training for 400 epochs with labels in Figure~\ref{fig:latent_2}. The samples have the same seed for the latents.
\end{document}